\documentclass[11pt]{article}
\usepackage[preprint]{acl}
\usepackage{times}
\usepackage{latexsym}
\usepackage{iftex}
\ifPDFTeX
\usepackage[T1]{fontenc}
\usepackage[utf8]{inputenc}
\usepackage{CJKutf8}
\newcommand{\target}{\begin{CJK*}{UTF8}{gbsn}不是X，而是Y\end{CJK*}}
\newcommand{\zh}[1]{\begin{CJK*}{UTF8}{gbsn}#1\end{CJK*}}
\else
\usepackage{fontspec,xeCJK}
\newcommand{\target}{不是X，而是Y}
\newcommand{\zh}[1]{#1}
\fi
\usepackage{amsmath,amssymb,booktabs,graphicx,array}
\usepackage{url}
\graphicspath{{figure/}}

\title{Not \textit{X}, but When: Diagnosing a Mandarin LLM Writing Tic
\\ through Contextual Appropriateness}
\author{%
\textbf{Yihang Peng}\textsuperscript{1}\thanks{Equal contribution.}%
\hspace{0.6em}%
\textbf{Yi Jing}\textsuperscript{1}\footnotemark[1]%
\hspace{0.6em}%
\textbf{Dewen Liu}\textsuperscript{2}%
\hspace{0.6em}%
\textbf{Weihan Meng}\textsuperscript{1}%
\hspace{0.6em}%
\textbf{Ying Liu}\textsuperscript{1}\thanks{Corresponding author.}%
\\
\textsuperscript{1}Tsinghua University
\hspace{1.3em}
\textsuperscript{2}Fudan University
\\
\texttt{yingliu@tsinghua.edu.cn}%
}
\begin{document}\maketitle
\begin{abstract}
Recurring expressions in model-written text are often treated as signatures of machine authorship, yet most of them serve legitimate functions. Whether such an expression is a tic depends on whether the model knows when it is called for. This judgment of appropriateness, central to communicative competence, remains largely unexamined for language models. We study it through the Mandarin corrective construction \emph{bushi X, ershi Y} (`not X, but Y'). Drawing on Gricean pragmatics and corrective negation, we introduce a decision tree that classifies uses as substantive, optional, or defective by testing whether the context makes X a live alternative and whether rejecting X is informative. Five commercial models use the construction 2.5 to 9.2 times as densely as a human reference. Of 307 annotated uses, only 1.30\% are defective while 52.8\% are optional, that is, correct but unneeded, so models fall short in appropriateness far more often than in correctness. Several reward models favor the construction even over quality-matched variants, an LLM judge favors it even where the context does not license it, and both supervised fine-tuning and preference optimization tend to raise its frequency in open-weight models. We hope this work encourages the development of models attentive to the appropriateness of their expressions.
\end{abstract}

\section{Introduction}
\label{sec:intro}
Large language models can now resolve real-world issues and complete multi-step tasks with tools \citep{jimenez2024swebench,yao2024taubench}, and a growing share of the text people read is machine-generated. In \citeposs{grice1975logic} account of cooperative conversation, a contribution should be as informative as the exchange requires and no more, and it should be relevant to the purpose at hand. An entirely accurate response can still fall short of these maxims by elaborating beyond what was asked or by addressing a question that was never at issue. Conformity to these maxims is therefore part of model quality alongside factual accuracy.

\begin{figure}[!t]
\centering
\includegraphics[width=\columnwidth]{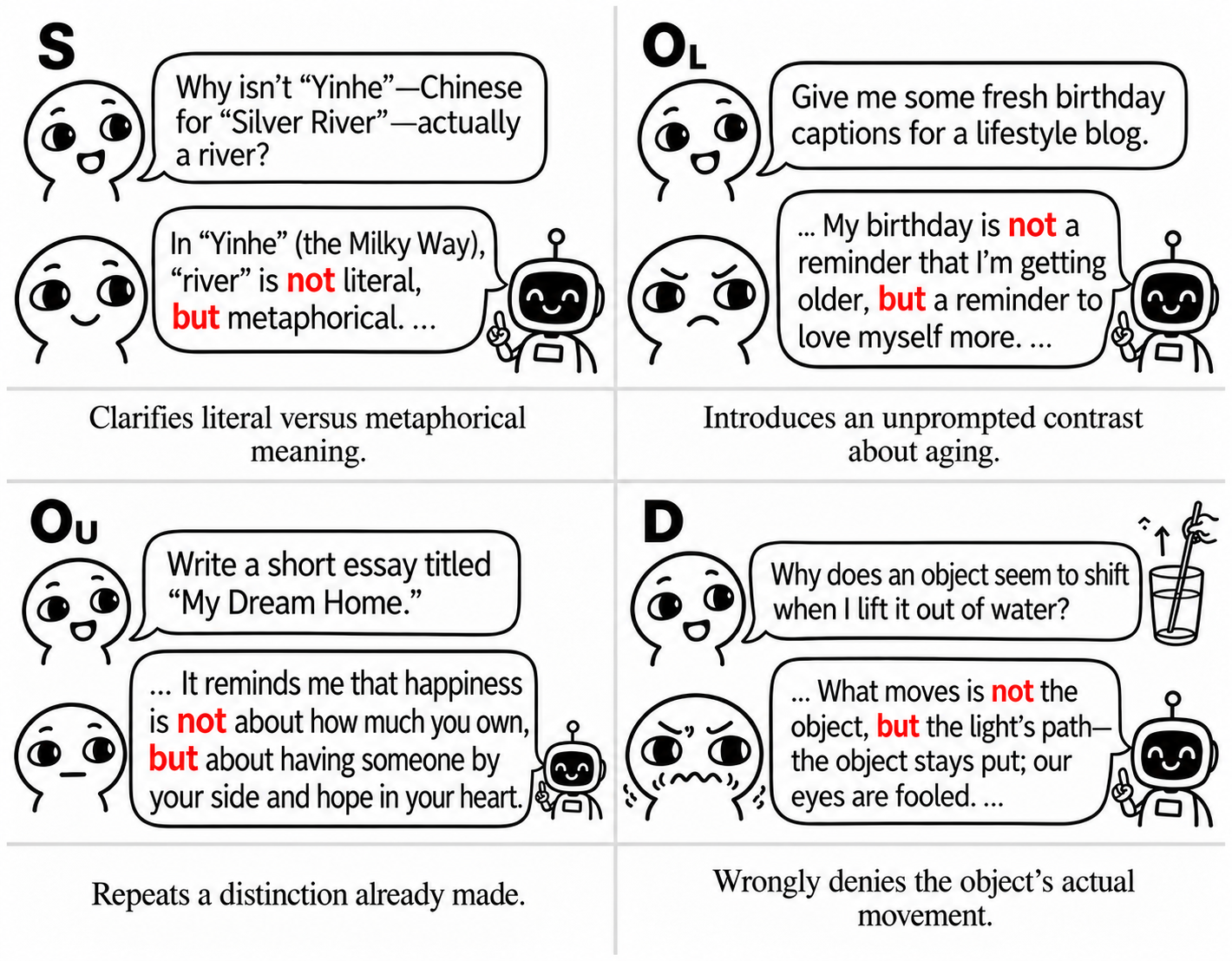}
\caption{One form, four discourse functions: substantive ($S$), unlicensed ($O_L$), redundant ($O_U$), and defective ($D$) uses of \emph{not X, but Y}. Translated excerpts retain the target sentence; ellipses mark omitted context.}
\label{fig:teaser}
\end{figure}

Model-written text also contains expressions that recur often enough for readers to treat them as signatures of machine authorship \citep{juzek-ward-2025-chatgpt,shaib2025slop}. Most of these expressions are ordinary resources of the language, and human writers use the same devices where the discourse calls for them. Whether a recurring expression amounts to a tic thus depends on whether the model knows when it is called for. \citet{hymes1972communicative} placed this knowledge at the core of communicative competence, as a judgment of whether an utterance is appropriate to its context, separate from whether it is grammatically possible and from how often it is in fact produced.

Research on language models, however, has mostly approached pragmatics from the listener's side, testing whether models recover implicatures and presuppositions \citep{ruis2023goldilocks,hu-etal-2023-fine,sravanthi-etal-2024-pub}, and recent comparisons show that competence in comprehension does not transfer to appropriate production \citep{sieker2026hypocritical}. When model production is studied, the usual measure is frequency, with words and phrases that are over-represented relative to human text treated as targets for suppression \citep{juzek-ward-2025-chatgpt,paech2025antislop}. Frequency answers the last of Hymes's questions, how often an expression is in fact produced. For expressions with a legitimate function, it conflates appropriate and inappropriate uses, and suppression removes both. Whether models command appropriateness itself remains to be examined.

We examine this question through corrective negation, which denies one alternative and asserts another in its place. Its English form, \emph{it's not X, it's Y}, is regularly cited as a hallmark of chatbot writing \citep{oremus2026atlantic}, and Chinese online commentary singles out the Mandarin \emph{bushi X, ershi Y} (`not X, but Y'; \target{}) in the same way, chiefly because the rejected X often appears to be a position that no one has taken.\footnote{For example, \url{https://blog.darkthread.net/blog/ai-slop-not-but-pattern/}.} Corrective negation is also the ordinary way to correct a misconception. Each use of the construction thus raises the two questions behind Grice's maxims (Figure~\ref{fig:teaser}): whether the context made X a live alternative (relevance), and whether rejecting X contributes information not already established (quantity). We ask whether models, like competent speakers, know when the construction is called for.

Linguistic theory makes these questions precise as conditions on licensing. As a construction, \emph{bushi X, ershi Y} conventionally pairs its form with a corrective function \citep{goldberg2006constructions,shao2010bushi,zuo2025opposition}. The correction presupposes alternatives that contrast along a shared dimension \citep{rooth1992focus}, and X need not have been asserted by anyone as long as it is accessible in the common ground \citep{toosarvandani2013corrective,stalnaker2002common}. What is rejected may be a proposition or the manner in which it is expressed, as in metalinguistic negation \citep{horn1985negation}, and whether the correction is informative depends on the question under discussion \citep{roberts2012information}. These accounts separate three cases that are formally identical: a correction that resolves a live alternative, one that is well formed but unneeded, and one that is defective because it rejects something true or misattributes a view.

We introduce a diagnostic framework built around an ordered decision tree (Figure~\ref{fig:framework}) that sorts each use of the construction into these three cases, first screening out defective uses and then testing the relevance and quantity conditions; a use that passes both is substantive, and one that fails either is optional. We apply the tree to answers from five commercial models on Chinese prompts and compare their usage with a human reference. We then investigate the sources of this preference, scoring each answer against minimally edited counterparts with reward models and an LLM judge, and tracing the construction's frequency across training stages of open-weight models and in public preference data.

For a substantial share of uses, models do not. They use the construction 2.5 to 9.2 times as densely as the human reference, and the shortfall lies in appropriateness more than in correctness, with only 1.30\% of annotated uses defective and 52.8\% classified as optional. Several reward models favor the construction even when variants match in general quality, and an LLM judge favors it even where the context does not license it. Along the training pipelines of open-weight models, both supervised fine-tuning and preference optimization tend to raise its frequency.

Our contributions are as follows:
\begin{itemize}\setlength{\itemsep}{0pt}
\item We formalize stylistic tics in LLM-generated text as a failure of appropriateness in the sense of communicative competence.
\item We distinguish uses that are correct from uses that are needed, and operationalize this distinction as a decision tree derived from linguistic theory.
\item We provide a method for diagnosing how reward models and training stages shape the preference for such expressions.
\end{itemize}

\section{Related Work}
\label{sec:related}
Our work draws on three lines of research. Studies of language models as speakers identify which expressions models overproduce; linguistic accounts of corrective negation specify when the construction is appropriate; and work on evaluator bias suggests where a preference for it may originate.

\subsection{Language Models as Speakers}
\label{sec:rw-speakers}
Research on language models has examined their pragmatic competence and their characteristic output largely in isolation. Pragmatic evaluations have mostly taken the listener's side, testing whether models recover implicatures and presuppositions \citep{ruis2023goldilocks,hu-etal-2023-fine,sravanthi-etal-2024-pub} or recognize constructional meaning \citep{weissweiler-etal-2022-better}. The speaker's side has a longer history in generation, where Gricean maxims constrain referring expressions \citep{dale1995computational} and pragmatic speakers reason about a listener in reference games \citep{frank2012predicting,andreas-klein-2016-reasoning}, though in tasks whose communicative goal is fixed in advance. Studies of open-ended model writing instead focus on its characteristic expressions, including over-represented words \citep{juzek-ward-2025-chatgpt,kobak2025delving}, recurring syntactic templates \citep{shaib-etal-2024-detection}, and broader measures of low-quality ``slop'' \citep{shaib2025slop}, with suppression proposed as a remedy \citep{paech2025antislop}. These studies establish which expressions models produce too often, which is a matter of frequency.

\subsection{Corrective Negation}
\label{sec:rw-negation}
Corrective negation is well described in both English and Mandarin, and these descriptions specify when a speaker uses it appropriately. The construction denies one alternative and asserts another in its place. Work on English shows that it presupposes alternatives contrasting along a shared dimension and can target the form of an utterance as well as its content \citep{horn1985negation,toosarvandani2013corrective}. Studies of \emph{bushi}\ldots\emph{ershi} describe the same corrective function in Mandarin. \citet{shao2010bushi} identify a pseudo-denial use, in which the negation does not retract any claim held in the context; \citet{tan2016bushi} compares the construction with English \emph{not A but B} and finds both semantic and metalinguistic correction in each; and \citet{zuo2025opposition} analyzes \emph{ershi} as the connective that lexicalizes correction, distinct from those for contrast and adversative relations. We turn these descriptions into conditions for assessing each use in model output.

\subsection{Evaluator Bias}
\label{sec:rw-evaluators}
The evaluators used to train and assess language models are biased toward surface features of responses, and training can carry these biases into model output. Reward models and LLM judges favor longer responses \citep{singhal2024long}, list formatting \citep{zhang-etal-2025-lists}, and style more broadly \citep{wu-aji-2025-style}, and LLM judges show position and verbosity biases \citep{zheng2023judging}. Preference optimization also narrows the range of model outputs \citep{kirk2024understanding}, and open post-training pipelines that release intermediate checkpoints make it possible to trace such changes stage by stage \citep{lambert2024tulu,olmo2025furious}. Unlike length or formatting, a corrective construction's appropriateness depends on context; the question is whether evaluators and training favor it only where it is called for.

\begin{figure*}[!t]
  \centering
  \includegraphics[width=\textwidth]{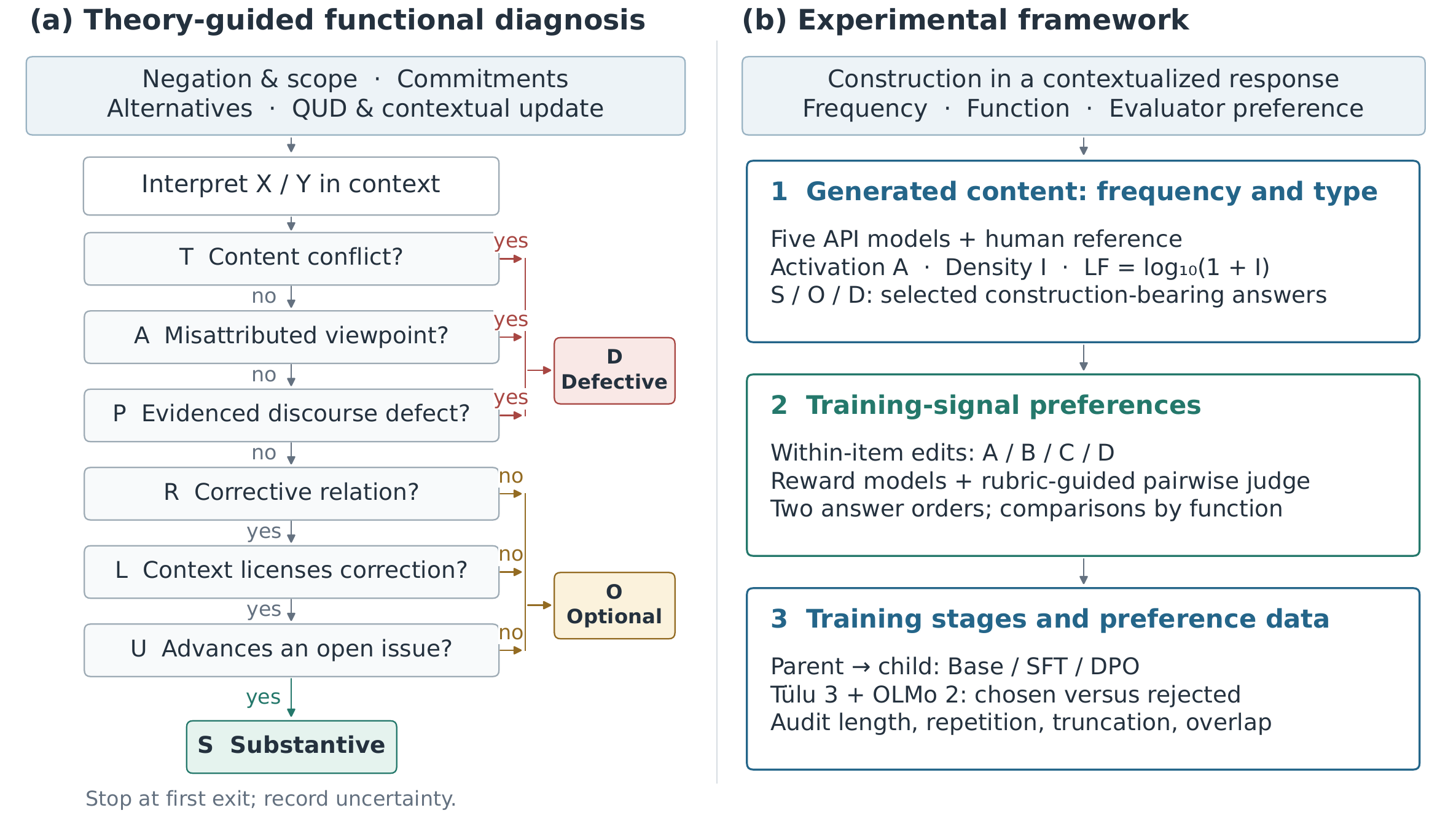}
  \captionsetup{hypcap=false}
  \caption{Diagnostic tree and experimental framework. (a) Recover the contextual reading, then test defects (T/A/P) and corrective function (R/L/U). (b) Connect generation patterns to evaluator preferences, checkpoint trajectories, and preference-data composition.}
  \label{fig:framework}
\end{figure*}

\section{Methods}
\label{sec:methods}
We propose a diagnostic framework that assesses each use of the construction by whether its context calls for it, rather than by how often it occurs. Its core is a decision tree, derived from corrective negation and Gricean pragmatics, that labels each occurrence as substantive, optional, or defective. We pair these labels with frequency measures, evaluator comparisons, and checkpoint analyses to trace the sources of the preference.

\subsection{Analytical Criteria}
We operationalize contextual appropriateness through corrective meaning and discourse contribution. A descriptive reading approximates $\neg p_X\land p_Y$; metalinguistic negation rejects wording or degree \citep{horn1985negation}, while reframing replaces an explanatory perspective. Alternatives share a recoverable contrast dimension \citep{rooth1992focus,toosarvandani2013corrective}, and their availability differs from a participant's endorsement \citep{stalnaker2002common}. Grice's Relation and Quantity maxims motivate separate tests of contextual licensing and informative contribution; Manner motivates checking interpretability \citep{grice1975logic}. We assess contribution against the current question under discussion (QUD), which tracks the distinctions still open in the discourse \citep{roberts2012information}. For instance, \emph{the bottleneck is not the CPU, but disk I/O} is licensed and informative when the user reports that upgrading the CPU did not help; it is unlicensed when nothing has raised the CPU, and redundant when the answer has already ruled the CPU out.

\subsection{An Evidence-Based Diagnostic Tree}
\label{sec:tree}
The unit is one occurrence in its complete answer and available context. Annotators recover X/Y, reading, scope, commitment holder, and QUD, then follow Figure~\ref{fig:framework}, citing evidence at each visited node.

\paragraph{Detect established defects: T $\rightarrow$ A $\rightarrow$ P.}
\textbf{T} tests for factual conflict, contradiction, or invalid exclusion under the recovered reading. \textbf{A} tests whether the correction misattributes a position. \textbf{P} tests for inappropriate rejection or disrupted interpretation, requiring a target signal, an independent defect basis, and a specific consequence, such as obstructed reference or unjustified issue replacement. Missing motivation alone is assessed at L. The first positive finding assigns \textbf{D (defective)}.

\paragraph{Assess corrective function: R $\rightarrow$ L $\rightarrow$ U.}
For remaining occurrences, \textbf{R} asks whether X/Y express a recoverable rejection and replacement. \textbf{L} requires a concrete contextual cue motivating the exclusion of X, independently of the contrast itself. \textbf{U} asks whether the correction advances an open distinction, supplies a missing replacement, or repairs a commitment under the QUD. Restating a completed correction fails U. The first negative answer assigns \textbf{O (optional)}, subtyped $O_R$, $O_L$, or $O_U$; passing all three assigns \textbf{S (substantive)}.

Annotators receive the original response without edited variants, evaluator scores, source identity, or prior labels. Factual status and confidence are recorded separately.

\subsection{Frequency and Function Measurement}
\label{sec:measurement}
For valid answer $i$, $c_i$ counts detected occurrences and $\ell_i^{(u)}$ counts Han characters ($u=H$) or all Unicode characters, including punctuation and whitespace ($u=C$). Activation, intensity, and log frequency are
\begin{align}
A&=N^{-1}\sum_i\mathbf{1}[c_i>0],\\
I_u&=10^4\frac{\sum_i c_i}{\sum_i\ell_i^{(u)}},\quad LF_u=\log_{10}(1+I_u).
\end{align}
Human--model comparisons average $I_H$ equally over four shared broad dimensions before computing $LF_H$. Full generation and checkpoint analyses average $A_b,I_C$ equally over four benchmarks; there $I$ abbreviates $I_C$. Empty or missing answers are excluded. Within a corpus, $I=10^4AB/\bar\ell$, where $B$ is occurrences per activated answer and $\bar\ell$ is mean answer length. This decomposition distinguishes activation from repetition and exposure. A fixed literal detector identifies candidates (Appendix~\ref{app:measurement}); functional proportions describe selected construction-bearing responses.

\subsection{Experimental Framework}
\label{sec:framework}
Generation analysis measures frequency and applies the tree. Evaluator experiments compare four variants: A retains \emph{not X, but Y}; B presents \emph{Y, not X}; C separates Y and the rejection across sentences; D retains only Y. A--B probes order, B--C sentence realization, and A--D deletes the rejected alternative and its corrective content.

For reward model $m$, $d_{im}^{AB}=(R_{imA}-R_{imB})/s_{mA}$ standardizes the contrast by original-arm reward variation. We also report A wins conditional on non-ties. Each pairwise judge evaluates two independent presentation orders. Restored-arm verdicts map to $1,.75,.5,.25,0$ in favor of A; their mean $p_i$ is an ordinal preference score. Functional sensitivity, $FS_G=\bar p_S-\bar p_G$, compares substantive uses with unlicensed uses ($G=O_L$) or with all optional uses ($G=O$); for reward models, $p$ is replaced by the standardized contrast $d$. A positive value means that the evaluator favors A more when the correction is needed. All intervals for $FS$ are 95\% item-bootstrap intervals.

Quality matching uses general-quality scores from HY3 \citep{tencent2026hy3}, retaining unchanged A--B/A--D texts scored equally on every applicable dimension except construction clarity. Checkpoint comparisons follow documented parent--child relations; preference-data audits compare context-matched chosen/rejected answers and examine length, language composition, and repetition.

\section{Experiments}
\label{sec:experiments}
Our experiments address two questions: whether models use the construction where it is called for, and where their preference for it comes from. We answer the first by comparing model and human usage in frequency and function, and the second by examining evaluator preferences, training stages, and public preference data.

\subsection{Frequency and Function}
\label{sec:generated}
\paragraph{Comparison.}
We examine 5,285 answers from five API models---GPT, DeepSeek, Gemini, GLM, and Qwen---to 1,061 Chinese prompts drawn from AlignBench \citep{liu-etal-2024-alignbench}, WritingBench \citep{wu2025writingbench}, COIG-CQIA \citep{bai2024coig}, and Arena-Hard-v2 \citep{arenahard2026repository}. A human reference comprises community answers, examination responses, student essays, and role-play dialogue. We compare construction density per 10,000 Han characters, averaging equally over four shared task dimensions: professional ability, general question answering, Chinese understanding, and writing. Tables~\ref{tab:human} and~\ref{tab:generation} summarize the human corpus and model generation panel; Appendix~\ref{app:measurement} provides the detection details.

\begin{figure}[!t]
\centering\includegraphics[width=\columnwidth]{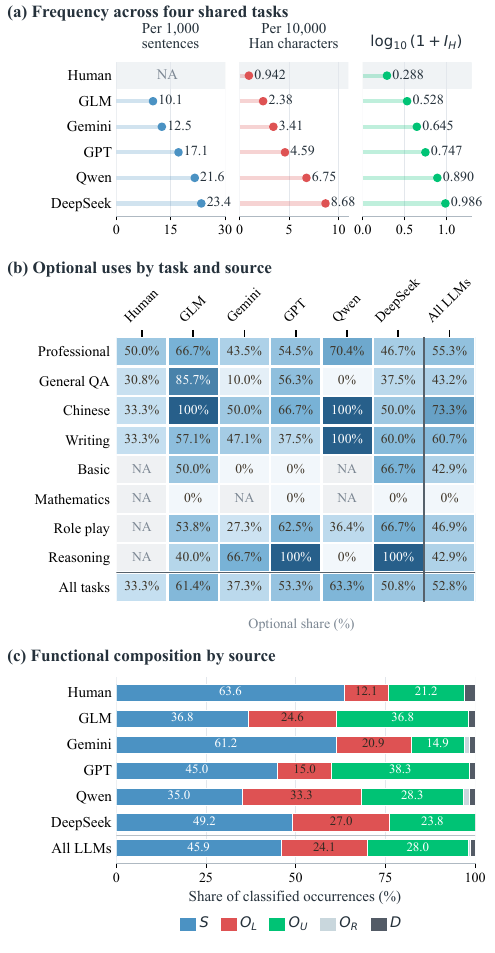}
\caption{\textbf{Frequency and corrective function.} (a) Occurrences per 1,000 punctuation-delimited sentences, per 10,000 Han characters ($I_H$), and $\log_{10}(1+I_H)$. Rates average equally over four shared tasks; the log is taken after averaging. Human sentence counts are unavailable (NA). (b) Optional-use percentages by task and source; NA denotes no classified cases. All LLMs pools model cases. (c) Functional shares: substantive ($S$), unlicensed ($O_L$), redundant ($O_U$), relation failure ($O_R$), and defective ($D$). Panels (b,c) include 307 selected model occurrences across four benchmarks and 33 eligible human-authored occurrences annotated with the same tree. Their denominators include every classified case; these descriptive samples differ from the full-answer frequency panel in (a).}
\label{fig:frequency}
\end{figure}

\paragraph{Models repeatedly select the construction.}
\label{sec:human}
Across the five models, density is 2.5--9.2 times the human reference (Figure~\ref{fig:frequency}a). The human reference answers different prompts, and its detector also admits comma-free and cross-sentence forms that the model detector excludes; this detection difference, if anything, understates the gap. The construction is thus a recurrent choice across models, rather than a peculiarity of one system. Its local contribution is a separate question: frequent corrections may resolve an outstanding alternative, but they may also introduce a contrast the answer does not need.

\paragraph{Many corrections lack a substantive corrective function.}
Of 308 selected occurrences, one cybersecurity-related item triggered a refusal from the Claude annotation service and yielded no valid label, leaving 307 occurrences for functional analysis in their full context. Labels were generated with \texttt{claude-sonnet-5-5} and subsequently audited. Optional uses account for 52.8\% of the 307 annotated cases (Figure~\ref{fig:frequency}b,c; Table~\ref{tab:functional-labels}). Contextual licensing or issue advancement accounts for 96.4\% of non-substantive decisions. The recurring problem is thus a correction introduced without a contextual reason, or repeated after its work has already been done. Only four occurrences are classified as defective.

\paragraph{Human-authored uses are more often substantive.}
We apply the same tree to the human reference. Of its 38 detected occurrences, five are quoted or metalinguistic and are excluded before tree application. Among the remaining 33, 63.6\% are substantive and 33.3\% optional, compared with 45.9\% and 52.8\% in model answers (Figure~\ref{fig:frequency}c). The difference is concentrated in unlicensed uses (12.1\% vs.\ 24.1\%); redundant uses differ less (21.2\% vs.\ 28.0\%). Models therefore not only use the construction more densely, but a larger share of their uses introduces an alternative that the context gives no reason to reject (Appendix~\ref{app:labels}).

\subsection{Evaluator Preference}
\label{sec:signals}
\paragraph{Two comparisons test evaluative preference.}
For each selected answer, we compare the original \emph{not X, but Y} with a \emph{reordered} version, \emph{Y, not X}, and an \emph{affirmative-only} version retaining Y. Reordering preserves the rejected alternative; affirmative-only editing removes it. Five reward models score full question--answer pairs. Claude Sonnet 5.5 and GPT-6.1-sol apply the same pairwise rubric in both presentation orders, yielding mean ordinal preference scores from 0 to 1, with .5 indicating a tie. Three of GPT's 1,204 planned judgments were blocked by the provider's content filter on one item. General-quality scores provide a further comparison among variants receiving identical scores on every applicable dimension (Appendix~\ref{app:evaluators}).

\begin{figure}[!t]
\centering\includegraphics[width=\columnwidth]{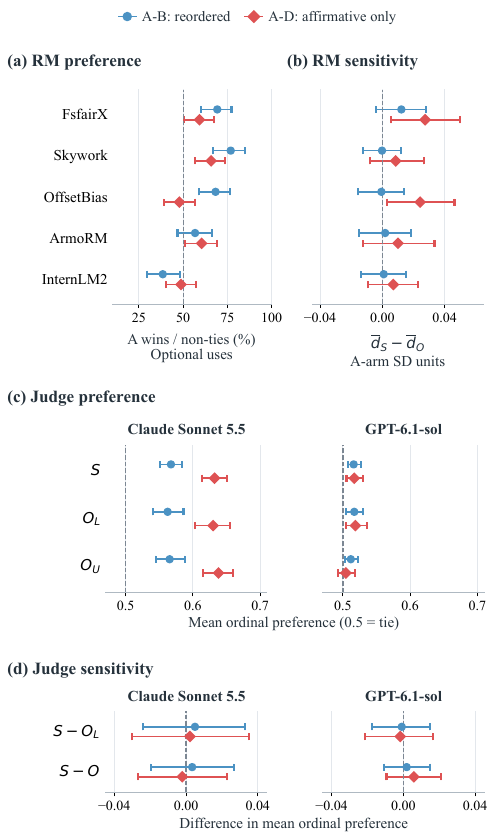}
\caption{\textbf{Evaluator preference and functional sensitivity.} A is the original, B reorders the correction, and D retains only Y. (a) RM non-tie A-win rates for optional uses. (b) Substantive-minus-optional reward contrasts, standardized by A-arm score SD. (c) Mean ordinal preference by function, separately for Claude Sonnet 5.5 and GPT-6.1-sol. (d) Their $FS_G=\bar p_S-\bar p_G$, with $O$ pooling $O_L$ and $O_U$. Dashed lines mark neutrality (a,c) or zero sensitivity (b,d). Panels (c,d) use each judge's complete pairs and shared scales across judges. Error bars are pointwise 95\% Wilson intervals (a) or 10,000 item-bootstrap intervals (b--d).}
\label{fig:signals}
\end{figure}

\paragraph{Reward preference extends to optional corrections.}
FsfairX \citep{dong2024rlhf}, Skywork \citep{liu2025skywork}, and OffsetBias \citep{park-etal-2024-offsetbias} prefer the original order in 68.3--76.8\% of non-tied optional-use comparisons (Figure~\ref{fig:signals}a). ArmoRM \citep{wang-etal-2024-interpretable} shows a weaker preference, whereas InternLM2 \citep{cai2024internlm2} favors the reordered version. The result identifies a recurring preference in three evaluators: they reward the original realization even where the tree classifies the correction as optional.

\paragraph{Judges favor retaining unlicensed corrections.}
For corrections without contextual licensing, Claude Sonnet 5.5 prefers the original over the affirmative-only answer (mean score 0.630) and over reordering (0.563; Figure~\ref{fig:signals}c; Table~\ref{tab:judge}). GPT-6.1-sol shows weaker but directionally consistent preferences (0.519 and 0.517, respectively). Critically, neither judge's preference varies with corrective function: substantive-minus-unlicensed differences are 0.002 and 0.005 for Claude, and $-0.002$ and $-0.001$ for GPT; all four 95\% intervals include zero (Figure~\ref{fig:signals}d). This insensitivity indicates that both judges reward the construction's surface form regardless of whether the context licenses it.

\paragraph{Ordering preference persists under quality-score matching.}
To examine whether reward models’ preference for the original order is associated with general answer quality, we retain original–reordered pairs with identical scores on every applicable quality dimension. Excluding reward-model ties, FsfairX, Skywork, and OffsetBias still prefer “not X, but Y” over “Y, not X” in 73.1--78.3\% of comparisons (Table~\ref{tab:quality}). Thus, these three reward models exhibit an ordering preference even when the quality scores do not distinguish the two versions.

\subsection{Training Stages and Data Attribution}
\label{sec:stages}
We examine SFT and DPO through 12 approximately 7B--8B checkpoints from DeepSeekMath \citep{shao2024deepseekmath}, Llama-3.1/T\"ulu-3 \citep{grattafiori2024llama3,lambert2024tulu}, OLMo-2 \citep{olmo2025furious}, MiMo \citep{xiaomi2025mimo}, and InternLM2 \citep{cai2024internlm2}. All five families provide Base and SFT checkpoints; T\"ulu and OLMo also provide DPO checkpoints. The panel yields 12,518 valid answers to the same 1,061 prompts used for the API models. Appendix~\ref{app:stages} provides the generation protocol.

\begin{figure*}[t]
\centering\includegraphics[width=\textwidth]{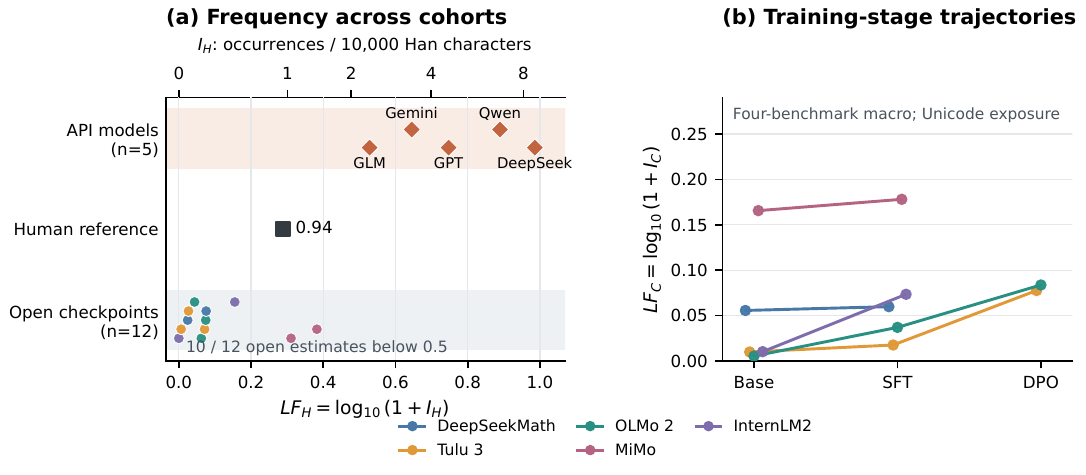}
\caption{\textbf{Cohort gap and SFT/DPO trajectories.} (a) Twelve open checkpoints from five families, the human reference, and five API models on four shared task dimensions. Points show $LF_H$, with $I_H$ on the upper axis; open points are offset vertically for visibility. (b) Descriptive Base--SFT--DPO trajectories using four-benchmark macro $LF_C$. Points use valid answers at each checkpoint; lines connect available stages. Nonempty truncated answers are retained.}
\label{fig:stages}
\end{figure*}

\begin{table}[!b]
\centering\footnotesize
\setlength{\tabcolsep}{3pt}
\begin{tabular}{@{}>{\raggedright\arraybackslash}p{0.12\columnwidth}>{\raggedright\arraybackslash}p{0.405\columnwidth}>{\raggedright\arraybackslash}p{0.405\columnwidth}@{}}
\toprule
& Swimmer's victory speech & Commentary on Messi's title \\
\midrule
Target & \zh{此刻我脑海里闪过的，不是终点触壁的那一瞬间，而是过去这几年的每一个清晨……} `not the moment I touched the wall, but every morning of these past years' & \zh{梅西的夺冠，不是因为他终于赢了，而是因为他用二十多年……换来了这一次最完整的救赎。} `Messi's title is not because he finally won, but because he traded twenty-odd years \ldots{} for this most complete redemption' \\
\addlinespace
Tree & $S$: winning makes the finish accessible; injury motivates rejecting it. & $O_L$ at L: no one reduces the title to winning; the target introduces X itself. \\
\addlinespace
Scores & RMs 3/3, 3/3; judge 0.75, 0.75 & RMs 2/3, 1/3; judge 0.75, 0.625 \\
\addlinespace
Reason & `rejects first, then asserts; more natural, more emotional impact' & `slightly odd logically but reads smoothly and passionately' \\
\bottomrule
\end{tabular}
\caption{Same evaluation, different function. Scores list A--B, then A--D. RMs counts how many of FsfairX, Skywork, and OffsetBias score the original higher; judge scores average two presentation orders (0.5 is a tie). Reasons are quoted from the judge's A--B output.}
\label{tab:case}
\end{table}

\paragraph{Open checkpoints use the construction less frequently than API models.}
Across the four shared task dimensions, open-checkpoint density ranges from 0 to 1.41 occurrences per 10,000 Han characters, compared with 0.942 for the human reference and 2.38--8.68 for the API models (Figure~\ref{fig:stages}a). Ten of the twelve open checkpoints fall below 0.500. The high-frequency pattern is therefore concentrated in the API cohort rather than shared by the selected open checkpoints.

\paragraph{Density rises across the displayed SFT and DPO stages.}
Under the four-benchmark Unicode measure, all five families have higher construction density at SFT than at Base. Both available DPO checkpoints also exceed their SFT predecessors, reaching the highest density within the displayed trajectories (Figure~\ref{fig:stages}b). Paired density changes on common successful prompts are also positive for all seven transitions (Table~\ref{tab:stage-effects}). These descriptive patterns associate SFT and DPO with increased use of the construction in this checkpoint panel.

\paragraph{Preferred training answers more often contain the construction.}
In public preference data, chosen answers activate the construction more often than rejected answers in both datasets: +14.7\% relative activation in T\"ulu 3 and +26.5\% in OLMo 2 (Table~\ref{tab:preference-data}). The direction is consistent across datasets and the gap is larger in OLMo 2. Both Han-content subsets in each corpus also show higher chosen-answer density. This enrichment is directionally consistent with the displayed DPO increases. Its magnitude depends on corpus composition and repetition.

\subsection{Case Study}
\label{sec:case}
Table~\ref{tab:case} presents two sports answers differing mainly in whether context supplies X. In a victory speech by a swimmer long troubled by injury, the finish is what a new champion would naturally recall, and the injury motivates setting it aside. In commentary on Messi's World Cup title, no one suggests that the title amounts merely to winning; the target raises this view only to reject it, although winning the final is precisely what the title is.

Yet the LLM judge favors the original in every judgment for both answers except one tie. For Messi, it concedes that the sentence is ``slightly odd logically'' and still prefers it because it ``reads smoothly and passionately.'' FsfairX and Skywork also score the original above the reordered version in both cases. The evaluators reward the construction's rhetorical effect, not whether the context raises the rejected alternative.

\section{Discussion and Conclusion}
Frequency and appropriateness are separate properties of model usage. Models produce the construction far more often than people do, and many of these uses are correct but not called for. Evaluators do not supply the missing judgment: several reward models prefer the construction after general-quality matching, and both judges prefer it within unlicensed uses. In the available checkpoints, its density also rises through SFT and DPO. Suppressing the form would remove its appropriate uses along with the rest; reducing tics of this kind requires evaluation and training signals that register whether an expression is called for. The diagnostic is not specific to Mandarin: screening for defects and then testing contextual licensing and issue advancement can be applied to English \emph{it's not X, it's Y} and to other expressions that are legitimate in some contexts and overused in others.

A controlled follow-up can test this attribution by holding SFT initialization, source composition, update budget, and prompts fixed, intervening on construction-bearing preference pairs with matched edits, and repeating training across seeds. Measuring activation, repetition, and local function would then separate changes in how often the construction is selected from changes in what it contributes.

\section*{Limitations}
\paragraph{Data coverage.}
Our analysis covers five commercial models, a limited set of task domains, and 307 annotated uses. Broader models, domains, and human data would strengthen the estimates.

\paragraph{Observational training evidence.}
The checkpoint and preference-data comparisons are observational, so they show that frequency rises along training pipelines but do not establish which stage or data causes the rise. Controlled retraining is needed to test this attribution.

\paragraph{Reader perception.}
We assess appropriateness through linguistic criteria and do not measure how readers perceive the construction. Whether unneeded uses cause annoyance or affect comprehension requires a separate user study.

\paragraph{Languages and constructions.}
We study a single Mandarin construction. Although the decision tree is designed to transfer to English \emph{it's not X, it's Y} and other recurring expressions, we have not applied it beyond \emph{bushi X, ershi Y}.

\section*{Ethical Considerations}
This section discusses the ethical considerations and broader impact of this work.

\paragraph{Potential Risks.}
The decision tree describes when a corrective construction is called for, and this description could be used to make model-written text resemble human writing more closely and thus harder to identify. Conversely, the frequency measures could be misread as evidence of machine authorship and applied to judge human writers, who also use the construction. Our analyses concern aggregate usage across models and corpora; they are not designed or validated for attributing individual texts to humans or machines.

\paragraph{Intellectual Property.}
The benchmarks, open-weight checkpoints, reward models, and preference mixtures used in this work are publicly released for research, and we use them in accordance with their respective licenses. Commercial models are accessed through their official APIs under the providers' terms of service. All materials are used solely for scientific research.

\paragraph{Data Privacy.}
The human reference is drawn from publicly available datasets, and personally identifiable information was removed before analysis. Model answers are generated from public benchmark prompts. The human annotators are students with a background in linguistics who participated voluntarily; no personal information about them is collected beyond their annotations.

\paragraph{Intended Use.}
The diagnostic framework is intended for analyzing contextual appropriateness in model-generated text and for examining how evaluators and training stages shape such usage. It is not intended for authorship detection or for evaluating individual writers.

\paragraph{Documentation of Artifacts.}
We will release our code and the annotation guidelines for the decision tree. Both are documented with respect to their scope: the Mandarin construction \emph{bushi X, ershi Y}, the task domains covered, and the procedures described in Appendices~\ref{app:measurement}--\ref{app:implementation}.

\paragraph{AI Assistants in Research or Writing.}
We use Cursor for code development assistance and a large language model for refining and polishing the language of the paper. The use of language models as annotators and judges is part of the method and is described in Section~\ref{sec:methods} and Appendices~\ref{app:labels} and~\ref{app:evaluators}.

\bibliography{paper_references}
\clearpage
\appendix
\section{Measurement and Annotation}
\label{app:measurement}
\paragraph{Human-reference sources.}
The human baseline draws on \href{https://github.com/zhangsheng93/cMedQA2}{cMedQA2} for medical community question--answer pairs; \href{https://github.com/Hello-SimpleAI/chatgpt-comparison-detection}{HC3} for Chinese human answers on finance, law, psychology, and everyday topics; and \href{https://github.com/BAAI-Zlab/COIG}{COIG} for Chinese-language examination answers and explanations. The \href{https://github.com/cnunlp/Chinese-Essay-Dataset-For-Pre-Training}{Chinese Essay Dataset} supplies student essays in six genres, and \href{https://aclanthology.org/2020.tacl-1.19/}{CrossWOZ} supplies travel-assistant dialogues role-played by human participants. We also use worked solutions for the 2016--2018 management entrance examinations hosted by \href{https://mba.upc.edu.cn/2024/0723/c19311a258464/page.htm}{China University of Petroleum (East China)}, solutions for the 2020 examination hosted by \href{https://mem.shnu.edu.cn/aa/78/c33053a830072/page.htm}{Shanghai Normal University}, and reference answers for the 2020 independent-admission mathematics examination hosted by the \href{https://fz.ahnu.edu.cn/info/1054/18825.htm}{High School Affiliated to Anhui Normal University}.

\paragraph{Corpora and generation.}
The human reference contains 572,412 Han characters and 38 occurrences across eight dimensions. The main comparison gives equal weight to the four dimensions represented by these occurrences: professional ability, general question answering, Chinese understanding, and writing. Han exposure counts code points in \texttt{U+4E00--U+9FFF}. The four-dimension mean is .942 occurrences per 10,000 Han characters; pooling all eight human dimensions gives .664. Human materials allow comma-free and cross-sentence realizations, whereas the model detector requires a comma.

The model panel contains 683 AlignBench, 100 WritingBench, 200 COIG-CQIA, and 78 Arena-Hard-v2 prompts. The five logged model identifiers are \texttt{gpt-5.6-luna}, \texttt{deepseek-v4.1-flash}, \texttt{gemini-3.8-flash}, \texttt{glm-5.3-flash}, and \texttt{qwen3.8-flash}. Each prompt is a single user message; endpoints retain default reasoning behavior and receive scene-dependent temperatures when supported. Of 5,305 expected answers, 5,285 are valid.

\begin{table*}[tp]
\centering\small
\setlength{\tabcolsep}{5pt}
\begin{tabular}{@{}lrrrr@{}}
\toprule
Task dimension & Han chars. & Count & $I_H$ & $LF_H$ \\
\midrule
\multicolumn{5}{@{}l}{\textbf{Four shared task dimensions (main comparison)}} \\
Professional tasks & 98,450 & 2 & 0.203 & 0.080 \\
General question answering & 100,020 & 13 & 1.30 & 0.362 \\
Chinese understanding & 100,146 & 11 & 1.10 & 0.322 \\
Writing & 102,823 & 12 & 1.17 & 0.336 \\
\textbf{Macro average of four shared dimensions} & --- & --- & \textbf{0.942} & --- \\
\addlinespace
\multicolumn{5}{@{}l}{\emph{Other human dimensions}} \\
Basic tasks & 44,867 & 0 & 0 & 0 \\
Logical reasoning & 17,683 & 0 & 0 & 0 \\
Mathematical calculation & 8,249 & 0 & 0 & 0 \\
Role play & 100,174 & 0 & 0 & 0 \\
\midrule
Corpus pooled (all 8 dimensions) & 572,412 & 38 & 0.664 & 0.221 \\
\bottomrule
\end{tabular}
\caption{Human corpus reference. $I_H$ counts occurrences per 10,000 Han characters; $LF_H=\log_{10}(1+I_H)$. The main comparison in the paper uses the macro average of the four shared dimensions ($I_H=0.942$). The corpus-pooled estimate over all eight dimensions is $I_H=0.664$. Dimensions with zero counts are shown for completeness; they are not used to compute the main comparison baseline.}
\label{tab:human}
\end{table*}

\begin{table*}[tp]
\centering\small
\setlength{\tabcolsep}{4pt}
\begin{tabular}{@{}lrrrrrrr@{}}
\toprule
 & \multicolumn{5}{c}{Four-benchmark generation panel} & \multicolumn{2}{c}{Four shared task dimensions} \\
\cmidrule(lr){2-6}\cmidrule(l){7-8}
Model & $N$ & $A$ (\%) & $I$ & 95\% CI for $I$ & $LF$ & $I_H$ & Model/Human \\
\midrule
GPT & 1061 & 31.6 & 2.54 & [2.15, 2.94] & 0.549 & 4.59 & $4.87\times$ \\
DeepSeek & 1058 & 23.5 & 2.97 & [2.49, 3.51] & 0.599 & 8.68 & $9.21\times$ \\
Gemini & 1061 & 17.0 & 1.20 & [0.987, 1.41] & 0.343 & 3.41 & $3.62\times$ \\
GLM & 1046 & 11.8 & 1.10 & [0.840, 1.38] & 0.321 & 2.38 & $2.52\times$ \\
Qwen & 1059 & 34.2 & 3.13 & [2.60, 3.75] & 0.616 & 6.75 & $7.17\times$ \\
\midrule
\emph{Human} & --- & --- & --- & --- & --- & \emph{0.942} & $1.00\times$ \\
\bottomrule
\end{tabular}
\caption{Five-model generation panel. $N$ is the number of valid responses. Within each benchmark, $A$ is the percentage of valid answers containing at least one occurrence, and $I$ is occurrences per 10,000 Unicode characters; the reported $A$ and $I$ average equally over four benchmarks. $LF=\log_{10}(1+I)$. The two rightmost columns use four shared task dimensions: $I_H$ is macro-average density per 10,000 Han characters, and Model/Human divides model $I_H$ by the unrounded human baseline (0.942 when rounded; Table~\ref{tab:human}). These ratios correspond to Figure~\ref{fig:frequency}a. Model names abbreviate the logged identifiers in the text.}
\label{tab:generation}
\end{table*}

\paragraph{Detection and frequency measures.}
The fixed detector finds non-overlapping literal matches, permits an optional preceding \emph{bing}, requires a Chinese or ASCII comma before \emph{ershi}, and limits the captured alternatives to 160 and 220 characters. Slots exclude sentence-final punctuation, semicolons, and newlines; intervening whitespace may span a newline. Nested starts and comma-free forms can be missed, and interrogative strings can produce spurious detections.

For valid answer $i$, let $c_i$ count occurrences and $\ell_i^{(u)}$ count Han or all Unicode characters. Activation and density are
\[
A=N^{-1}\sum_i\mathbf{1}[c_i>0],
\qquad I_u=10^4\frac{\sum_i c_i}{\sum_i\ell_i^{(u)}}.
\]
For Figure~\ref{fig:frequency}a, sentence density is $10^3\sum_i c_i/\sum_i s_i$, where $s_i$ counts punctuation-delimited units. We strip line-initial numbered-list prefixes, split at Chinese full stops and Chinese/ASCII question and exclamation marks, and split at ASCII periods unless flanked by digits. Segments must contain a Unicode letter; a final unterminated segment is retained. Newlines and semicolons do not split units. Sentence and Han densities are computed within each of the four shared tasks and then averaged equally. Human sentence counts are unavailable. The log-frequency panel transforms the macro Han density after averaging.

The archived log scale is $LF_u=\log_{10}(1+I_u)$. Human comparisons average Han density over four task dimensions. Benchmark analyses average activation and Unicode density equally over four benchmarks. Within a corpus, density decomposes into activation, occurrences per activated answer, and mean answer length.

Full-panel uncertainty uses 5,000 prompt-cluster bootstraps stratified by benchmark. Between-model tests use common successful prompts and within-prompt exchanges, with Holm correction over ten activation and ten density contrasts separately.

\subsection{Functional Annotation and Human Comparison}
\label{app:labels}
The sample comprises 308 selected response groups. One cybersecurity-related item triggered a refusal from the Claude annotation service and yielded no valid label; the functional analysis therefore covers 307 annotated targets. Labels were generated with \texttt{claude-sonnet-5-5} and subsequently audited. Annotation excludes templates, purely metalinguistic quoted strings, and known incomplete materials. Annotators assess the original response without edited variants or evaluator scores, recover the supported reading and question under discussion, and record factual uncertainty separately. An implicit question under discussion requires contextual evidence independent of the target contrast.

The annotated sample contains 141 substantive, 162 optional, and four defective uses (Table~\ref{tab:functional-labels}). Optional uses comprise 52.8\% of all 307 annotated cases and 53.5\% of the 303 non-defective cases. Figure~\ref{fig:frequency}b,c includes all classified cases. Of the 166 non-substantive uses, 160 (96.4\%) terminate at contextual licensing or issue advancement.

\begin{table}[tbp]
\centering\small
\setlength{\tabcolsep}{4pt}
\begin{tabular}{@{}lrr@{}}
\toprule
Outcome / nested subtype & Count & Share (\%) \\
\midrule
\textbf{1. Defective ($D$)} & \textbf{4} & \textbf{1.30} \\
\multicolumn{3}{@{}l}{\emph{First positive check: $T\rightarrow A\rightarrow P$}} \\
\quad $T$: Truth conflict & 2 & 0.651 \\
\quad $A$: Misattribution & 2 & 0.651 \\
\quad $P$: Pragmatic defect & 0 & 0 \\
\midrule
\textbf{2. Optional ($O$)} & \textbf{162} & \textbf{52.8} \\
\multicolumn{3}{@{}l}{\emph{If no defect, first failure: $R\rightarrow L\rightarrow U$}} \\
\quad $R$: Relation failure & 2 & 0.651 \\
\quad $L$: No contextual licensing & 74 & 24.1 \\
\quad $U$: No issue advancement & 86 & 28.0 \\
\midrule
\textbf{3. Substantive ($S$)} & \textbf{141} & \textbf{45.9} \\
\multicolumn{3}{@{}l}{\emph{No defect; all $R/L/U$ checks passed}} \\
\midrule
Total & 307 & 100 \\
\bottomrule
\end{tabular}
\caption{Functional composition of the 307 annotated cases. S/O/D are top-level outcomes; R/L/U (optional) and T/A/P (defective) are mutually exclusive subcategories assigned in priority order (R $\rightarrow$ L $\rightarrow$ U; T $\rightarrow$ A $\rightarrow$ P). One cybersecurity-related item was excluded due to an annotation refusal. Labels were generated by an LLM and reviewed by humans; inter-rater agreement is reported in Table~\ref{tab:human-agreement}. This table describes the current sample ($N=307$) and should not be interpreted as a precise estimate of population proportions.}
\label{tab:functional-labels}
\end{table}

\paragraph{Agreement with the human annotations.} 
We compare annotations from two independent human annotators, both with a background in linguistics, on the same 100 cases (Table~\ref{tab:human-agreement}). We report exact agreement and unweighted Cohen's $\kappa$ for the top-level outcomes (S/O/D) and the seven terminal labels (S, R, L, U, T, A, P). Terminal-label agreement requires the full classification, including the subtype when applicable, to match. Both annotators labeled the cases independently without consensus discussion. The $\kappa$ values indicate moderate agreement \citep{landis1977measurement}. The expected drop at the terminal level reflects the inherent difficulty of distinguishing fine-grained pragmatic subcategories (e.g., $O_L$ vs.\ $O_U$), even for trained linguists.

\begin{table}[tbp]
\centering\small
\setlength{\tabcolsep}{4pt}
\begin{tabular}{@{}lrrrr@{}}
\toprule
Label level & $N$ & Matches & Agreement & $\kappa$ \\
\midrule
Top-level ($S/O/D$) & 100 & 77 & 77.0\% & 0.577 \\
Terminal (7 labels) & 100 & 66 & 66.0\% & 0.521 \\
\bottomrule
\end{tabular}
\caption{Inter-annotator agreement on 100 cases. Two independent annotators with linguistics backgrounds labeled all cases. Exact Matches (n) and Exact Agreement (\%) report raw agreement. $\kappa$ is unweighted Cohen's $\kappa$. The terminal level includes the seven labels (S, R, L, U, T, A, P). The moderate $\kappa$ values reflect the inherent difficulty of fine-grained pragmatic classification, particularly distinguishing lack of contextual licensing ($O_L$) from no discourse advancement ($O_U$).}
\label{tab:human-agreement}
\end{table}

\paragraph{Human-authored functional baseline.}
The human corpus contains 38 detected occurrences. After excluding five quoted or metalinguistic instances before tree application, Codex classifies 33 eligible occurrences under the same protocol: 21 substantive, four unlicensed, seven redundant, and one defective. These labels did not undergo the review applied to the model sample. Section~\ref{sec:generated} compares them with the model sample. They are automated annotations of human-authored text and are distinct from the human raters described above.

\paragraph{Discourse position.}
Target position is its first start index divided by answer length. Median positions are 0.341 for substantive corrections, 0.544 for unlicensed corrections, and 0.862 for redundant corrections. The substantive--unlicensed difference has $p<.001$. Summary cues occur before 7.80\%, 14.9\%, and 32.6\% of these targets. Both optional subtypes occur later than substantive corrections. These associations are descriptive and do not enter the evaluator tests.

\section{Evaluator Design and Supplementary Results}
\label{app:evaluators}
\paragraph{Variants and reward models.}
Each selected response yields four variants. For compatibility with the released data, A denotes the original, B the reordered version, C a cross-sentence version, and D the affirmative-only version. The main text uses the first, second, and fourth. The cross-sentence variant is not used in the reported comparisons.

ArmoRM-8B, FsfairX-8B, InternLM2-7B-Reward, OffsetBias-8B, and Skywork-Reward-V2-Qwen3-8B receive the question and full answer without functional or variant labels. Overlength inputs are excluded. The first, second, and fourth scorers cover 252 complete groups; InternLM2 and Skywork cover 305. Their common intersection contains 252 groups. The scoring release preserves all originals and repairs 67 edited variants in 35 groups.

Mean reward contrasts are standardized by the original-answer score standard deviation. Functional-sensitivity intervals use 10,000 item bootstraps.

\paragraph{General-quality matching.}
The numerical-quality rubric contains 24 general dimensions. Matching requires unchanged scored texts and equal scores on every applicable dimension, excluding the construction-clarity manipulation check. Repaired variants whose scores refer to earlier texts are excluded. Repeated scores have 78.5\% exact agreement and 96.2\% agreement within one point; weighted $\kappa=.822$. Table~\ref{tab:quality} reports the retained groups, wins, losses, and ties. Matching does not condition on functional category.

\begin{table}[tbp]\centering\small\setlength{\tabcolsep}{3pt}
\begin{tabular}{llrrr}\toprule
Pair & RM & $N$ & W/L/T & Rate\\\midrule
A--B & FsfairX & 57 & 36/10/11 & 0.783\\
A--B & Skywork & 67 & 30/11/26 & 0.732\\
A--B & OffsetBias & 57 & 38/14/5 & 0.731\\
A--B & ArmoRM & 57 & 24/20/13 & 0.545\\
A--B & InternLM2 & 67 & 16/28/23 & 0.364\\
A--D & FsfairX & 59 & 39/18/2 & 0.684\\
A--D & Skywork & 71 & 36/16/19 & 0.692\\
A--D & OffsetBias & 59 & 29/26/4 & 0.527\\
A--D & ArmoRM & 59 & 38/17/4 & 0.691\\
A--D & InternLM2 & 71 & 28/24/19 & 0.538\\
\bottomrule\end{tabular}
\caption{General-quality score-matched subset on unchanged scored texts. W/L/T are A wins, losses, and ties; rate is W/(W+L), not W/$N$. Matching requires equality on every applicable general-quality dimension.}
\label{tab:quality}\end{table}

\paragraph{Pairwise judgment.}
The Arena-Hard rubric assesses correctness, helpfulness, relevance, concision, completeness, and applicable creativity. The logged judge aliases are \texttt{claude-sonnet-5-5} and \texttt{gpt-6.1-sol}. Each judge generates a reference answer and independently evaluates both presentation orders. After restoring variant identity, strong/weak original preference, tie, and weak/strong counterpart preference map to $1,.75,.5,.25,0$. The two-order average is an ordinal preference score. Twelve GPT judgments were recovered with the original 8,192-token client-side output cap omitted; prompts, reasoning effort, and scoring were unchanged.

There are 301 eligible groups: 141 substantive, 74 unlicensed, and 86 redundant. Defective, relation-failure, and unresolved cases are excluded. Claude has 297 complete two-order pairs per comparison: 139/138 substantive, 74/73 unlicensed, and 84/86 redundant for A--B/A--D. GPT has 300 per comparison, comprising 141 substantive, 74 unlicensed, and 85 redundant pairs for each edit. Figure~\ref{fig:signals} reports non-tie reward-model win rates with Wilson 95\% intervals. Judge preference and functional-sensitivity intervals use 10,000 item bootstraps; sensitivity resamples substantive and optional groups independently.

\begin{table*}[tp]
\centering\small
\setlength{\tabcolsep}{5pt}
\begin{tabular}{llrrr}
\toprule
Contrast & Group & $n/N$ & Mean score & 95\% CI \\
\midrule
A--D & Overall & 297/301 & 0.633 & [0.621, 0.646] \\
A--D & S & 138/141 & 0.632 & [0.613, 0.650] \\
A--D & $O_L$ & 73/74 & 0.630 & [0.603, 0.656] \\
A--D & $O_U$ & 86/86 & 0.638 & [0.615, 0.661] \\
A--B & Overall & 297/301 & 0.566 & [0.554, 0.577] \\
A--B & S & 139/141 & 0.567 & [0.551, 0.585] \\
A--B & $O_L$ & 74/74 & 0.563 & [0.541, 0.586] \\
A--B & $O_U$ & 84/86 & 0.565 & [0.545, 0.588] \\
\bottomrule
\end{tabular}
\caption{Claude Sonnet 5.5 rubric-guided pairwise judgments by functional group. Scores average two presentation orders after restoring variant identity; values above 0.5 favor A. Intervals use 10,000 item bootstraps. $n/N$ denotes complete pairs/eligible groups. Overall covers the 301 substantive, unlicensed, and redundant cases; the two relation failures and four defective cases are excluded.}
\label{tab:judge}
\end{table*}

\paragraph{Functional sensitivity of the judges.}
Mean preference for retaining unlicensed corrections over affirmative-only answers is 0.630 (95\% CI $[0.603,0.656]$) for Claude and 0.519 ($[0.505,0.535]$) for GPT. Their substantive-minus-unlicensed and substantive-minus-pooled-optional differences range from $-0.002$ to 0.006 across the two edits; all eight 95\% intervals include zero (Figure~\ref{fig:signals}d).

\paragraph{Functional differences across reward models.}
The pooled reward-model $FS_O$, based on standardized contrasts, is .00194 for reordering, with a 95\% interval of $[-.00924,.0125]$. For affirmative-only editing it is .0154, with an interval of $[.000439,.0301]$. The latter comparison removes corrective content as well as its realization. The main claim rests on positive preference within optional and unlicensed uses, rather than on an equivalence claim across functional groups.

\section{Checkpoint and Preference-Data Protocols}
\label{app:stages}
\paragraph{Checkpoint panel.}
The panel contains Base and SFT checkpoints from DeepSeekMath, Llama-3.1/T\"ulu-3, OLMo-2, MiMo, and InternLM2, plus DPO checkpoints from T\"ulu and OLMo: 12 checkpoints in total. All receive the same 1,061 prompts as the API panel, yielding 12,518 valid answers. Base checkpoints use a neutral question--answer frame; chat checkpoints use native templates without an added system message.

Decoding uses temperature .1 for 725 prompts and .7 for 336 according to scene, with $\mathrm{top}\_p=1$. The context cap is the smaller of the native model limit and 32,768 tokens; the initial generation budget is the smaller of 16,384 tokens and remaining context. Empty answers and overlength-input records are excluded. Nonempty truncated responses are retained. Each plotted point uses the valid answers at that checkpoint and averages density equally over the four benchmarks. Table~\ref{tab:stage-effects} gives the corresponding descriptive paired changes on common successful prompts.

\begin{table*}[tp]
\centering\small
\setlength{\tabcolsep}{7pt}
\begin{tabular}{llrr}
\toprule
Family & Transition & Paired $N$ & $\Delta I_C$ \\
\midrule
DeepSeekMath & Base$\to$SFT & 1,028 & +0.0154 \\
Llama/T\"ulu 3 & Base$\to$SFT & 1,055 & +0.0187 \\
Llama/T\"ulu 3 & SFT$\to$DPO & 1,061 & +0.154 \\
OLMo 2 & Base$\to$SFT & 1,006 & +0.0767 \\
OLMo 2 & SFT$\to$DPO & 1,018 & +0.124 \\
MiMo & Base$\to$SFT & 1,053 & +0.131 \\
InternLM2 & Base$\to$SFT & 1,056 & +0.165 \\
\bottomrule
\end{tabular}
\caption{Descriptive parent--child density changes for SFT and DPO. $\Delta I_C$ is child minus parent density per 10,000 Unicode characters, averaged equally over four benchmarks. Each transition uses common successful prompts and retains nonempty truncated answers. These paired changes need not equal differences between the marginal checkpoint points in Figure~\ref{fig:stages}b.}
\label{tab:stage-effects}
\end{table*}

\begin{figure*}[tp]
\centering\includegraphics[width=\textwidth]{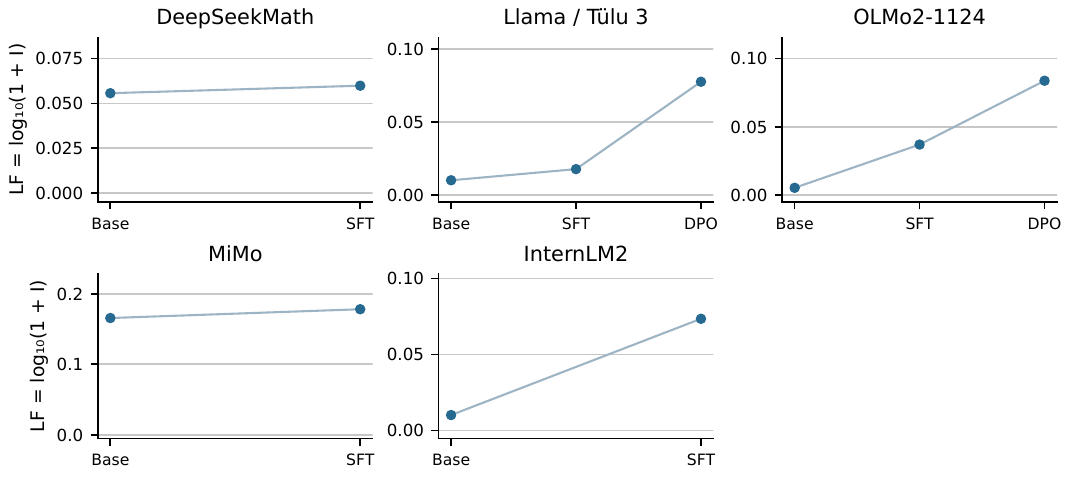}
\caption{Descriptive Base--SFT--DPO trajectories by family. Points show benchmark-macro log-frequency, $LF_C=\log_{10}(1+I_C)$, using valid answers at each checkpoint and retaining nonempty truncated answers. Lines connect available stages; vertical scales differ across panels.}
\label{fig:full-stages}
\end{figure*}

Density reflects completion length and repeated use as well as construction selection. Excluding truncated answers reverses the OLMo Base--SFT direction, while concentrated repetition contributes substantially to MiMo's density. The displayed trajectories describe the retained-output protocol.

\paragraph{Preference-data audit.}
We analyze the fixed T\"ulu-3-8B and OLMo-2-1124-7B preference mixtures. Of 651,239 released rows, 648,868 have valid context-matched chosen/rejected final assistant answers: 271,958 T\"ulu pairs and 376,910 OLMo pairs. The datasets share 264,594 complete prompts, covering 99.74\% and 72.34\% of their unique valid prompts. The audit uses only the final assistant answer with identical prior context. Table~\ref{tab:preference-data} reports construction activation in each answer arm.

\begin{table*}[tp]
\centering\small
\setlength{\tabcolsep}{4pt}
\begin{tabular}{@{}llrrrrrr@{}}
\toprule
Data & Arm & Pairs & Hits & Instances & $A$ (\%) & $A$ per 10k pairs & $\Delta_{\mathrm{rel}}$ vs.\ rejected \\
\midrule
T\"ulu 3 & Chosen & 271,958 & 86 & 145 & 0.03162 & 3.162 & +14.7\% \\
T\"ulu 3 & Rejected & 271,958 & 75 & 91 & 0.02758 & 2.758 & --- \\
OLMo 2 & Chosen & 376,910 & 253 & 307 & 0.06712 & 6.712 & +26.5\% \\
OLMo 2 & Rejected & 376,910 & 200 & 272 & 0.05306 & 5.306 & --- \\
\bottomrule
\end{tabular}
\caption{Public preference data, restricted to complete context-matched chosen/rejected pairs and final assistant responses. Hits count activated answers. $A$ per 10k pairs $=10^4\times\mathrm{Hits}/\mathrm{Pairs}$. $\Delta_{\mathrm{rel}}$ is the relative increase of chosen over rejected activation, computed from unrounded counts. Chosen answers consistently show higher activation than rejected answers in both datasets, with a larger gap in OLMo 2. These rates are corpus-pooled, not benchmark-macro estimates.}
\label{tab:preference-data}
\end{table*}

Preference-data intervals use a prompt-cluster-robust delta method.

For the Han-prompt proxy, the released prompt field (serialized when structured) must contain at least 20 Han characters and a Han-to-non-whitespace-character ratio of at least .2. For the both-Han-answers proxy, each chosen and rejected final answer must separately meet these thresholds. Subset intervals are unadjusted; density exposure remains all Unicode characters, including whitespace.

Paired conditional-logit models adjusting log length ratio and Han proportion give construction odds ratios of .947 for T\"ulu (95\% CI $[.673,1.332]$) and 1.040 for OLMo ($[.839,1.289]$). One T\"ulu chosen answer supplies 41 of 145 occurrences; removing its prompt cluster reduces the intensity difference by 73.8\%. OLMo chosen answers repeat the construction less within activated answers: 1.213 versus 1.360 instances.

\section{Implementation Details}
\label{app:implementation}
Local inference for the open-weight checkpoints and reward models used NVIDIA A800 and H800 GPUs, totaling 8.35 A800 GPU hours and 17.07 H800 GPU hours. Checkpoint generation used vLLM 0.11.0 \citep{kwon2023vllm}. Reward-model scoring used the Hugging Face Transformers 4.56.1 implementations \citep{wolf-etal-2020-transformers} with PyTorch 2.8.0 \citep{paszke2019pytorch} and CUDA 12.8. The five commercial models, the annotation model, and both pairwise judges were accessed through their providers' APIs. Decoding settings for the checkpoint panel are given in Appendix~\ref{app:stages}, and those for the API panel in Appendix~\ref{app:measurement}.

\end{document}